\documentclass[runningheads,a4paper]{llncs}

\usepackage{amssymb}
\usepackage{amsmath}

\makeatletter
\def\thanks#1{\protected@xdef\@thanks{\@thanks\protect\footnotetext{#1}}}
\makeatother

\usepackage{subfig}
\usepackage{caption}
\usepackage{algorithm} 
\usepackage[noend]{algpseudocode}  
\usepackage{graphicx}

\usepackage{url}
\newcommand{\keywords}[1]{\par\addvspace\baselineskip
\noindent\keywordname\enspace\ignorespaces#1}

\begin{document}
\mainmatter  % start of an individual contribution

\title{SANOM Results for OAEI 2019}


\titlerunning{SANOM results for OAEI 2019}

\author{\thanks{Copyright © 2019 for this paper by its authors. Use permitted under\hfill \break
Creative Commons License Attribution 4.0 International (CC BY 4.0).}Majid Mohammadi%
\and Amir Ahooye Atashin \and Wout Hofman\and Yao-Hua Tan
}

\institute{Faculty of Technology, Policy and Management, Delft University of Technology, The Netherlands,\\
TNO Research institute, The Netherlands.}

\toctitle{Lecture Notes in Computer Science}
\tocauthor{Authors' Instructions}
\maketitle

\begin{abstract}
Simulated annealing-based ontology matching (SANOM) participates for the second time at the ontology alignment evaluation initiative (OAEI) 2019. This paper contains the configuration of SANOM and its results on the anatomy and conference tracks. In comparison to the OAEI 2017, SANOM has improved significantly, and its results are competitive with the state-of-the-art systems. In particular, SANOM has the highest recall rate among the participated systems in the conference track, and is competitive with AML, the best performing system, in terms of F-measure. SANOM is also competitive with LogMap on the anatomy track, which is the best performing system in this track with no usage of particular biomedical background knowledge. SANOM has been adapted to the HOBBIT platfrom and is now available for the registered users.
\emph{abstract} environment.
\end{abstract}

%%%%%%%%%%%%%%%%%%%%%%%%%%%%%%%%%%%%%%%%%%%%%%%%
%%%%%%%%%%%%% Introduction
%%%%%%%%%%%%%%%%%%%%%%%%%%%%%%%%%%%%%%%%%%%%%%%%
\section{System Representation}
SANOM takes advantages of the well-known simulated annealing (SA) to discover the shared concepts between two given ontologies \cite{sanom}. A potential alignment is modeled as a state in the SA whose evolution would result in a more reliable matching between ontologies. The evolution requires a fitness function in order to gauge the goodness of the intermediate solutions to the ontology matching problem. 

A fitness function should utilize the lexical and structural similarity metrics to estimate the fineness of an alignment. The version of SANOM participated this year uses both lexical and structural similarity metrics, which are described in the following. 

%%%%%%%%%%%%%%%%%%%%%%%%%%%%%%%%%%%%%%%%%%%%%%%%%%%%%%%
\subsection{Lexical Similarity Metric}
The cleaning of strings before the similarity computation is essential to increase the chance of mapping entities. SANOM uses the following pre-processing techniques to this end:
\begin{itemize}
\item \textbf{Tokenization.} It is quite common that the The terminology of concepts are constructed from a bag of words (BoW). The words are often concatenated by white space, the capital letter of first letters, and several punctuations such as $"-"$ or $"\_"$. Therefore, they need to be broken into individual words and then the similarity is computed by comparing the bag of words together.

\item \textbf{Stop word removal.} Stop words are the typical words with no particular meaning. The stop words should be detected by searching the tokens (identified after tokenization) in a table containing all possible stop words. The Glasgow stop word list is utilized in the current implementation \footnote{http://ir.dcs.gla.ac.uk/resources/linguistic utils/stop words} . 

\item \textbf{Stemming.} Two entities from the given ontologies might refer to a similar concept, but they are named differently due to various verb tense, plural/singular, and so forth. Therefore, one needs to recover the normal words so that the similar concepts will have higher similarity. The Porter stemming method is used for this matter \cite{porter}.
\end{itemize}

After the pre-processing step, the strings of two concepts can be given to a similarity metric in order to calibrate the degree of similarity between concepts. The base similarity metric computes the sameness of tokens obtained from each entity. The current version of SANOM takes advantage of two similarity metrics and take their maximum as the final similarity of two given tokens. One of this similarity metric is for sole comparison of stirngs, and the other one is to guage the linguistic relation of two given names. These similarity metrics are:
\begin{itemize}
\item \textbf{Jaro-Winkler metric.} The combination of TF-IDF and Jaro-Winkler is popular and has been sucessful in ontology alignment as well. Similarly, SANOM uses Jaro-Winkler with the threshold $0.9$ as one of the base similarity metrics. 
%%%% this is copy
\item \textbf{WordNet-based metric.} The linguistic heterogeneity is also rampant in various domains. Therefore, the existence of a similarity metric to measure the lingual closeness of two entities is absolutely essential. In this study, the relatedness of two given tokens are computed by the Wu and Palmer measure \cite{wupalmer} and is used as a base similarity metric with the threshold 0.95.
\end{itemize}

%%%%%%%%%%%%%%%%%%%%%%%%%%%%%%%%%%%%%%%%%%%%%%%%%%%%%%%
%% THIS IS COPY
\subsection{Structural Similarity Metric}
The preceding string similarity metric gives a high score to the entities which have lexical or linguistic proximity. Another similarity of two entities could be derived from their positions in the given ontologies. 

We consider two structural similarity measures for the current implementation of SANOM:
\begin{itemize}
\item The first structural similarity is gauged by the subsumption relation of classes. If there are two classes $c_1$ and $c_2$ whose superclasses are $s_1$ and $s_2$ from two given ontologies $O_1$ and $O_2$, then the matching of classes $s_1$ and $s_2$ would increase the similarity of $c_1$ and $c_2$. Let $s$ be a correspondence mapping $s_1$ to $s_2$, then the increased similarity of $c_1$ and $c_2$ is gauged by
\begin{align}
f_{structural}(c_1,c_2) = f(s).
\end{align}

\item Another structural similarity is derived from the properties of the given ontologies. The alignment of two properties would tell us the fact that their corresponding domain and/or ranges are also identical. Similarly, if two properties have the analogous domain and/or range, then it is likely that they are similar as well. 

The names of properties and even their corresponding core concepts are not a reliable meter based on which they are declared a correspondence. A recent study has shown that the mapping of properties solely based on their names would result in high false positive and false negative rates, e.g. there are properties with identical names which are not semantically related while there are semantically relevant properties with totally distinct names.

The current implementation treats the object and data properties differently. For the object properties $op_1$ and $op_2$, their corresponding domains and ranges are computed as the concatenation of their set of ranges and domains, respectively. Then, the fitness of the names, domains, and ranges are computed by the Soft TF-IDF.  The final mapping of two properties is the average of top two fitness scores obtained by the Soft TF-IDF. For the data properties, the fitness is computed as the similarity average of names and their corresponding domain.

On the other flow of alignment, it is possible to derive if two classes are identical based on the properties. Let $e_1$ and $e_2$ be classes, $op_1$ and $op_2$ be the object properties, and $R_1$ and $R_2$ are the corresponding ranges, then the correspondence $c = (e_1,e_2)$ is evaluated as
\begin{align}
f_{structural}(c) = \frac{f_{string}(R_1,R_2) + f_{string}(op_1,op_2)}{2}.
\end{align}
\end{itemize}

%%%%%%%%%%%%%%%%%%%%%%%%%%%%%%%%%%%%%%%%%%%%%%%%
%%%%%%%%%%%%% Results
%%%%%%%%%%%%%%%%%%%%%%%%%%%%%%%%%%%%%%%%%%%%%%%%
\section{Results}
This section contains the results obtained by SANOM on the anatomy and conference track.

\subsection{Anatomy Track}
The anatomy track is one of the earliest benchmarks in the OAEI. The task is about aligning the Adult Mouse anatomy and a part of NCI thesaurus containing the anatomy of humans. Each of the ontologies has approximately 3,000 classes, which are designed carefully and are annotated in technical terms.

The best performing systems in this track use a biomedical background knowledge. Thus, their results are not comparable with SANOM which does not use any particular background knowledge. Among other systems, LogMap \cite{logmap} is best one with no use of a background knowledge.

Table \ref{tab:anatomy performance} tabulates the precision, recall, and F-measure of SANOM and LogMap on the anatomy track. According to this table, the recall of SANOM is slightly higher than LogMap which means that it could identify more correspondences than LogMap. However, the precision of LogMap is better than SANOM with the margin of three percent. The overall performance of SANOM is quite close to LogMap since their F-measure has only $1\%$ difference.

\begin{table}
\centering
\begin{tabular}{c|ccc}
System & Precision & F-measure & Recall \\ \hline
LogMap     & 0.918 & 0.88  & 0.846 \\
SANOM      & 0.888 & 0.87 & 0.853  \\
\end{tabular}
\caption{The precision, recall, and F-measure of SANOM and LogMap on the OAEI anatomy track.}
\label{tab:anatomy performance}
\end{table}

\begin{table}
    \centering
    \begin{tabular}{c|ccc|ccc|ccc}
& \multicolumn{3}{c}{SANOM} & \multicolumn{3}{c}{AML} &\multicolumn{3}{c}{LogMap} \\
& P&F&R & P&F&R & P&F&R \\ \hline
cmt-conference     & 0.61 & 0.74 & 0.93 & 0.67 & 0.59 & 0.53 & 0.73 & 0.62 & 0.53 \\
cmt-confOf         & 0.80 & 0.62 & 0.50 & 0.90 & 0.69 & 0.56 & 0.83 & 0.45 & 0.31 \\
cmt-edas           & 0.63 & 0.69 & 0.77 & 0.90 & 0.78 & 0.69 & 0.89 & 0.73 & 0.62 \\
cmt-ekaw           & 0.54 & 0.58 & 0.64 & 0.75 & 0.63 & 0.55 & 0.75 & 0.63 & 0.55 \\ 
cmt-iasted         & 0.67 & 0.80 & 1.00 & 0.80 & 0.89 & 1.00 & 0.80 & 0.89 & 1.00 \\
cmt-sigkdd         & 0.85 & 0.88 & 0.92 & 0.92 & 0.92 & 0.92 & 1.00 & 0.91 & 0.83 \\
conference-confOf  & 0.79 & 0.76 & 0.73 & 0.87 & 0.87 & 0.87 & 0.85 & 0.79 & 0.73 \\
conference-edas    & 0.67 & 0.74 & 0.82 & 0.73 & 0.69 & 0.65 & 0.85 & 0.73 & 0.65 \\
conference-ekaw    & 0.66 & 0.70 & 0.76 & 0.78 & 0.75 & 0.72 & 0.63 & 0.55 & 0.48 \\
conference-iasted  & 0.88 & 0.64 & 0.50 & 0.83 & 0.50 & 0.36 & 0.88 & 0.64 & 0.50 \\ 
conference-sigkdd  & 0.75 & 0.77 & 0.80 & 0.85 & 0.79 & 0.73 & 0.85 & 0.79 & 0.73 \\
confOf-edas        & 0.82 & 0.78 & 0.74 & 0.92 & 0.71 & 0.58 & 0.77 & 0.63 & 0.53 \\
confOf-ekaw        & 0.81 & 0.83 & 0.85 & 0.94 & 0.86 & 0.80 & 0.93 & 0.80 & 0.70 \\
confOf-iasted      & 0.71 & 0.63 & 0.56 & 0.80 & 0.57 & 0.44 & 1.00 & 0.62 & 0.44 \\
confOf-sigkdd      & 0.83 & 0.77 & 0.71 & 1.00 & 0.92 & 0.86 & 1.00 & 0.83 & 0.71 \\
edas-ekaw          & 0.71 & 0.72 & 0.74 & 0.79 & 0.59 & 0.48 & 0.75 & 0.62 & 0.52 \\
edas-iasted        & 0.69 & 0.56 & 0.47 & 0.82 & 0.60 & 0.47 & 0.88 & 0.52 & 0.37 \\
edas-sigkdd        & 0.80 & 0.64 & 0.53 & 1.00 & 0.80 & 0.67 & 0.88 & 0.61 & 0.47 \\
ekaw-iasted        & 0.70 & 0.70 & 0.70 & 0.88 & 0.78 & 0.70 & 0.75 & 0.67 & 0.60 \\
ekaw-sigkdd        & 0.89 & 0.80 & 0.73 & 0.80 & 0.76 & 0.73 & 0.86 & 0.67 & 0.55 \\
iasted-sigkdd      & 0.70 & 0.80 & 0.93 & 0.81 & 0.84 & 0.87 & 0.71 & 0.69 & 0.67 \\ \hline
Average    & 0.74 & 0.72 & 0.73 & 0.84 & 0.74 & 0.67 & 0.84 & 0.68 & 0.59 \\

    \end{tabular}
    \caption{The precision, recall, and F-measure of SANOM, AML, and LogMap on various datasets on the conference track}
    \label{tab:conference performance}
\end{table}

%%%%%%%%%%%%%%%%%%%%%%%%%%%%%%%%%%%%%%%%%%%%%%%%%%%%
\subsection{Conference Track}
The conference comprises the pairwise alignment of seven ontologies. Table \ref{tab:conference performance} displays the precision, recall, and F-measure of SANOM, LogMap, and AML \cite{aml} on the conference track. AML and LogMap are the top two systems in terms of precision and recall.

According to Table \ref{tab:conference performance}, the recall of SANOM is superior to both LogMap and AML. SANOM's average recall is $7\%$ and $14\%$ more than those of AML and LogMap, respectively, but its precision is $10\%$ less than both of the systems. Overall, the performance of SANOM is quite competitive with the top performing systems in the conference track.  
%%%%%%%%%%%%%%%%%%%%%%%%%%%%%%%%%%%%%%%%%%%%%%%%%%%%
\subsection{Large BioMed Track}
The conference comprises the pairwise alignment of seven ontologies. Table \ref{tab:largebio performance} displays the precision, recall, and F-measure of SANOM, LogMap, and AML \cite{aml} on the Large BioMed track. AML and LogMap are the top two systems in terms of precision and recall.

\begin{table}
\centering
    \begin{tabular}{c|ccc|ccc|ccc}
& \multicolumn{3}{c}{SANOM} & \multicolumn{3}{c}{AML} &\multicolumn{3}{c}{LogMap} \\
& P&F&R & P&F&R & P&F&R \\ \hline
FMA-NCI (whole)     & 0.61 & 0.74 & 0.841 & 0.805 & 0.59 & 0.881 & 0.856 & 0.831 & 0.808 \\
FMA-SNOMED (whole)  & 0.905 & 0.283 & 0.167 & 0.685 & 0.697 & 0.710 & 0.840 & 0.730 & 0.645 \\
SNOMED-NCI (whole)      & 0.868 & 0.618 & 0.479 & 0.862 & 0.765 & 0.687 & 0.867 & 0.706 & 0.596 \\
    \end{tabular}
    \caption{The precision, recall, and F-measure of SANOM, AML, and LogMap on various datasets on the Large BioMed track}
    \label{tab:largebio performance}
\end{table}
%%%%%%%%%%%%%%%%%%%%%%%%%%%%%%%%%%%%%%%%%%%%%%%%
%%%%%%%%%%%%% Conclusion
%%%%%%%%%%%%%%%%%%%%%%%%%%%%%%%%%%%%%%%%%%%%%%%%
\section{Conclusion}
SANOM only participated in the OAEI 2019 anatomy, conference and Large BioMed track. For the next year, we have aims to participate in more tracks so that the performance of SANOM can be compared with that of the state-of-the-art systems in other tracks as well. Another avenue to improve the system is to equip it with a proper biomedical background knowledge since most of the OAEI tracks are from this domain.

%\section*{Reference}
\bibliographystyle{plain}
\bibliography{main}

\begin{thebibliography}{1}

\bibitem{aml}
Daniel Faria, Catia Pesquita, Booma Balasubramani, Teemu Tervo, David
  Carri{\c{c}}o, Rodrigo Garrilha, Francisco Couto, and Isabel~F Cruz.
\newblock Results of aml participation in oaei 2018.
\newblock In {\em Proceedings of the 13th International Workshop on Ontology
  Matching co-located with the 17th International Semantic Web Conference},
  volume 2288, 2018.

\bibitem{logmap}
Ernesto Jim{\'e}nez-Ruiz and Bernardo~Cuenca Grau.
\newblock Logmap: Logic-based and scalable ontology matching.
\newblock In {\em International Semantic Web Conference}, pages 273--288.
  Springer, 2011.

\bibitem{sanom}
Majid Mohammadi, Wout Hofman, and Yaohua Tan.
\newblock Simulated annealing-based ontology matching.
\newblock 2018.

\bibitem{porter}
Martin~F Porter.
\newblock An algorithm for suffix stripping.
\newblock {\em Program}, 14(3):130--137, 1980.

\bibitem{wupalmer}
Zhibiao Wu and Martha Palmer.
\newblock Verbs semantics and lexical selection.
\newblock In {\em Proceedings of the 32nd annual meeting on Association for
  Computational Linguistics}, pages 133--138. Association for Computational
  Linguistics, 1994.

\end{thebibliography}

\end{document}